%% file: main.tex
\documentclass[conference]{IEEEtran}
\IEEEoverridecommandlockouts

\usepackage{cite}
\usepackage{amsmath,amssymb,amsfonts}
\usepackage{algorithmic}
\usepackage{graphicx}
\usepackage{textcomp}
\usepackage{xcolor}
\usepackage{multirow}
\usepackage{orcidlink}
\hypersetup{hidelinks}
\begin{document}

\title{Don't Just Impute, Detect: Task-Aware Diffusion Pipelines for Edge Anomaly Detection}

\author{\IEEEauthorblockN{Jeongsu Park\,\orcidlink{0009-0001-6846-1095},
Taehyeon Yun\,\orcidlink{0009-0000-2380-7008},
Geonwoo Kim\,\orcidlink{0009-0007-1567-9107},
Yuji Lim\,\orcidlink{0009-0009-1595-5532},
Moohong Min\,\orcidlink{0000-0001-8979-1344}$^{*}$}
\IEEEauthorblockA{\textit{Sungkyunkwan University}, Seoul, Republic of Korea}
\thanks{$^{*}$Corresponding authors.}
}

\maketitle

\begingroup
\renewcommand{\thefootnote}{}
\footnotetext{%
  \copyright{} 2026 IEEE.
  Personal use of this material is permitted.
  Permission from IEEE must be obtained for all other uses,
  in any current or future media, including
  reprinting/republishing this material for advertising
  or promotional purposes, creating new collective works,
  for resale or redistribution to servers or lists,
  or reuse of any copyrighted component of this work
  in other works.
}
\endgroup

\begin{abstract}
Industrial Internet of Things (IIoT) edge pipelines routinely lose contiguous blocks of multi-sensor data to network dropouts and hardware faults. The standard remedy is a disjointed ``impute-then-detect'' pipeline whose imputer is trained only for Mean Squared Error (MSE). Such an imputer fills a missing block with the most probable normal pattern: a partially observed anomaly is smoothed back toward normality, and boundary artifacts on normal data raise false alarms. We call this imputation bias and propose the Task-Aware Diffusion Refiner (TADR), a conditional diffusion refiner trained against the loss landscape of a frozen downstream detector. Because back-propagating through the sampling trajectory is intractable, the task-aware terms act on the one-step clean-window estimate that the noise prediction already yields, so training costs no more than plain diffusion training. Under strict Point-to-Point evaluation over five seeds with paired significance tests, TADR raises F1 by $0.338$ over the unrefined detector on SMD ($p = 0.0015$), and the anomaly enhancement objective alone contributes $0.463$ in ablation. We also characterise when the method helps: the gain tracks the continuous, mutually redundant channels available and the anomaly evidence the outage removes, and it is absent on telemetry datasets dominated by one-hot command channels and on PSM. At ten sampling steps TADR runs in $41.4$ ms per window on a single CPU core, sustaining about $24$ concurrent 1 Hz streams from a $4.03$ MB model. Code and per-seed results are available at \url{https://github.com/int-park422/TADR}.
\end{abstract}


\input{sections/sec1}

\input{sections/sec2}

\input{sections/sec3}

\input{sections/sec4}

\input{sections/sec5}

\input{sections/sec6}

\input{sections/sec7}

\section*{Acknowledgment}
This research was supported by Development of Field Technology for Responding to Illegal drugs Program through the Korea Institutes of Police Technology (KIPoT) funded by the Ministry of Science and ICT \& Korean National Police Agency (RS-2026-25535833).

\bibliographystyle{references/IEEEtran}
\bibliography{references/references}

\end{document}

%% file: sections/sec1.tex
\section{Introduction}
\label{sec1}

The Industrial Internet of Things (IIoT) and Cyber-Physical Systems (CPS) rely heavily on continuous multi-sensor data to monitor operational status and ensure structural integrity. However, real-world IIoT infrastructures often operate under resource-constrained edge conditions where hardware malfunctions and network dropouts frequently disrupt data collection. Unlike simple and isolated missing points, these communication failures typically result in block missing—continuous sequences of data loss across multiple timestamps and channels \cite{cao2018brits,miao2021generative}. This structural corruption not only poses a severe threat to downstream anomaly detection systems but also creates a critical processing bottleneck within the edge-to-cloud data pipeline. The same concern extends to security-critical monitoring such as on-site sensing for illicit drug detection, where the rare events of interest can coincide with lost data, and an imputer that fills the gap with normal-looking values erases precisely what the system exists to find.

To handle missing data, standard industrial pipelines typically employ a disjointed two-step "impute-then-detect" strategy \cite{luo2018multivariate,tashiro2021csdi}. In this paradigm, incomplete data is first reconstructed using generative imputation models optimized purely for Mean Squared Error (MSE), followed by the application of state-of-the-art anomaly detectors such as Anomaly Transformer \cite{xu2022anomaly} or TimesNet \cite{wu2023timesnet}. While this intuitively restores the structural integrity of the time series, it exposes a vulnerability in the pipeline architecture: \textit{imputation bias} \cite{zong2018deep,audibert2020usad,li2019mad}. Because the generative model is isolated from the downstream task, it defaults to synthesizing statistically probable normal patterns---suppressing the very attack signals the detector is looking for, and leaving boundary artifacts that inflate false alarms. Section \ref{sec3} formalises this failure mode.

To address the computational inefficiencies and architectural vulnerabilities of disjointed data preparation pipelines, we propose the Task-Aware Diffusion Refiner (TADR), an end-to-end data preprocessing optimization framework tailored for edge infrastructures facing severe block missing conditions. Rather than merely smoothing data to minimize reconstruction error, TADR fundamentally redefines the objective of the data preparation phase \cite{rasul2021autoregressive}. We introduce a task-aware diffusion mechanism where the generative refinement process is explicitly guided by the loss landscape of the downstream anomaly detector. By bridging the semantic gap between data reconstruction and anomaly identification, TADR refines incomplete sequences to sharpen critical anomaly peaks while preserving normal background patterns. Because the detector stays frozen, TADR is a plug-in module that requires no structural modification of the detector; we find it effective for detectors that score by reconstruction error.

The main contributions of this paper are as follows.

\begin{itemize}\setlength{\itemsep}{2pt}
\item We expose how the disjointed nature of standard reconstruction pipelines induces imputation bias under block missing conditions, squandering edge resources and inflating false alarms.

\item We propose TADR, a conditional diffusion refiner coupled to the frozen detector's objective. Because back-propagating through the sampling trajectory is intractable, the task-aware terms act on the one-step estimate the noise prediction already yields, leaving training no more costly than plain diffusion training.

\item Under strict Point-to-Point evaluation with five seeds, confidence intervals and paired tests, TADR raises P2P F1 by $+0.338$ over the unrefined detector; the enhancement objective alone accounts for $+0.463$ of the ablation gain, and the iterative formulation for a further $+0.300$ over a single-pass refiner.

\item Rather than claiming uniform superiority, we characterise \emph{when} task-aware refinement helps: the gain tracks the continuous, mutually redundant sensor signal available, and vanishes when channels are predominantly discrete or when the detector already recovers most anomalies unaided. Adversarial manipulation of the observable channels, poisoning, and concept drift are outside our evaluation (Section~\ref{sec6}).

\item We profile TADR at batch size one on a single CPU core rather than only on server GPUs, and identify the sampling budget at which quality is retained, establishing a deployable operating point.
\end{itemize}

%% file: sections/sec2.tex
\section{Related Work}
\label{sec2}

\subsection{Deep Learning for Time-Series Anomaly Detection}
Deep learning techniques have significantly advanced multivariate time-series anomaly detection. The field has transitioned from traditional Recurrent Neural Networks (RNNs) to architectures that better capture complex temporal and cross-channel dependencies \cite{malhotra2015long,park2018multimodal,su2019robust}. Transformer-based models, such as the Anomaly Transformer \cite{xu2022anomaly}, utilize anomaly attention mechanisms to distinguish abnormal patterns by measuring the discrepancy between series and prior associations. The inverted Transformer (iTransformer) extends this by embedding entire temporal sequences as tokens to model multivariate correlations \cite{liu2024itransformer}. Concurrently, convolutional methods like TimesNet \cite{wu2023timesnet} address multi-periodic variations through 2D convolutions, while KAN-AD explores the parameter efficiency of Kolmogorov-Arnold Networks for anomaly detection \cite{liu2024kan}. However, a shared limitation among these detectors is their inherent assumption of ideal network conditions and structural data integrity. Because they are designed for stable and continuous data streams, they experience severe performance degradation in practical IIoT edge environments where sensor malfunctions and network dropouts render the input sequences highly fragmented \cite{cao2018brits,tang2020joint}.

\subsection{Disjointed Pipelines and Imputation Bias}
To manage missing values in real-world deployments, the standard industrial practice employs a disjointed ``impute-then-detect'' pipeline. Early methods applied statistical interpolation or forward-filling, which fails to capture the dynamics of IIoT signals \cite{little2019statistical,che2018recurrent}; later approaches reconstruct missing segments with autoencoders or recurrent networks optimised for Mean Squared Error (MSE) \cite{yoon2018estimate,du2023saits}. Because such imputers are isolated from the detection objective and trained on normal operation, they are prone to imputation bias \cite{cao2018brits,du2023saits,zhao2020multivariate}, the failure mode we formalise in Section~\ref{sec3}. The resulting loss of precision has historically been masked by the Point Adjustment (PA) metric, which inflates recall regardless of false alarms \cite{li2019mad,audibert2020usad}.

\subsection{Generative Refinement and System-Level Optimization}
Denoising Diffusion Probabilistic Models (DDPMs) and score-based generative models are increasingly used for time-series imputation because of their high-fidelity generation \cite{ho2020denoising,song2020score}. CSDI trains a conditional diffusion model that exploits correlations among observed values \cite{tashiro2021csdi}, and TIMBA combines bidirectional Mamba blocks with a diffusion framework to capture long-range dependencies \cite{solis2024timba}. Both remain reconstruction-centric, optimised for conditional MSE \cite{rasul2021autoregressive,lin2024diffusion}, and diffusion imputers additionally pay for iterative sampling at inference. Whether that cost is justified for detection depends on what the reconstruction does to the detector's score, which their objective does not consider: under strict Point-to-Point (P2P) evaluation we find such imputers can help substantially or collapse precision (Section~\ref{sec5}). TADR keeps the diffusion architecture and sampler of CSDI but replaces the objective with one coupled to the frozen detector, so that the sampling cost buys detection quality rather than statistical smoothness alone.

%% file: sections/sec3.tex
\section{Motivation and Problem Formulation}
\label{sec3}

\begin{figure*}[t]
\centering
\includegraphics[width=0.75\textwidth]{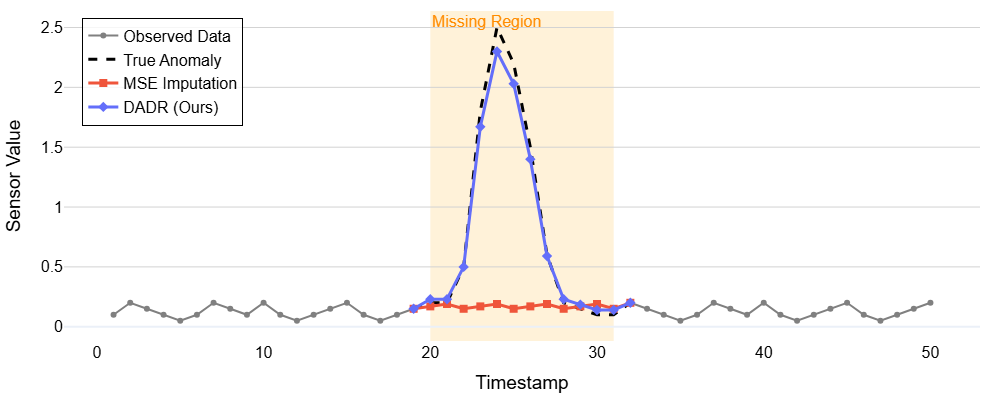}
\caption{Conceptual illustration of imputation bias and pipeline failure, demonstrated using a sampled anomaly region from the normalized Server Machine Dataset (SMD). When an anomaly coincides with a block missing event, standard MSE-based optimization wastes computational resources to synthesize a statistically probable normal pattern, inadvertently erasing the true anomaly. TADR resolves this by directly preserving critical sharp variations for the detector.}
\label{fig:imputation_bias}
\end{figure*}

\subsection{Problem Definition: Anomaly Detection under Block Missing}
In typical IIoT infrastructures, we monitor a multivariate time-series sequence \( \mathcal{X} = \{x_1, x_2, \dots, x_T\} \in \mathbb{R}^{T \times C} \), where \( T \) is the total number of timestamps and \( C \) is the number of sensor channels. Following standard anomaly detection practices, the continuous sequence is divided into deeply overlapped sliding windows \cite{garg2021evaluation}. Let \( X \in \mathbb{R}^{W \times C} \) denote a single sliding window of length \( W \).

Under ideal network conditions, the downstream anomaly detection model \( f_{\theta}(\cdot) \) maps the complete window \( X \) to an anomaly score vector \( S \in \mathbb{R}^W \), where a higher score indicates a higher probability of an anomalous event, such as a cyber-physical attack or equipment failure. However, physical edge environments often suffer from strict hardware constraints, network dropouts, and sensor malfunctions, leading to incomplete data matrices. We model this structural corruption using a binary missing mask \( M \in \{0, 1\}^{W \times C} \), where \( 0 \) indicates a missing value and \( 1 \) indicates an observed value. 

Unlike random missing values (Missing Completely At Random, MCAR) often assumed in general data science \cite{rubin1976inference}, sensor failures in harsh IIoT settings typically result in block missing scenarios \cite{yi2016st}. A block missing event of length \( L \) starting at timestamp \( t_{start} \) implies that \( M_{t, c} = 0 \) for all \( t \in [t_{start}, t_{start} + L - 1] \) across one or multiple channels. Consequently, the anomaly detector only receives the heavily corrupted input \( \tilde{X} = X \odot M \), where \( \odot \) denotes element-wise multiplication, which severely disrupts the temporal receptive field required for accurate detection. A systematic comparison of generative detectors under varied noise reports the same asymmetry: block-type perturbation degrades detection most sharply, and does so architecture-specifically \cite{park2025robustness}.

\subsection{Motivation: The Systemic Dilemma of Impute-then-Detect}
To counter the block missing problem and restore the receptive field, the conventional edge pipeline employs a pre-trained generative refiner \( g_{\phi}(\cdot) \) as a data preparation step before the detector. The goal of \( g_{\phi} \) is to reconstruct the original window \( X \) from the corrupted input \( \tilde{X} \). The reconstructed window \( \hat{X} = g_{\phi}(\tilde{X}) \) is then fed into the detector \( f_{\theta} \). 

Whether utilizing basic autoencoders or advanced generative architectures like CSDI and TIMBA, the refiner \( g_{\phi} \) is universally optimized using the Mean Squared Error (MSE) objective:
\[ \mathcal{L}_{\mathrm{MSE}} = \| X - g_{\phi}(\tilde{X}) \|_2^2 \]

While minimizing \( \mathcal{L}_{\mathrm{MSE}} \) reduces the statistical distance between the reconstructed and original signals, it leads to two pipeline failure modes. First, anomalies are rare, abrupt deviations from normal operation. Because the training data consist almost exclusively of normal operation, \( g_{\phi} \) learns the conditional expectation of the normal distribution, \( \mathbb{E}[X_{\mathrm{normal}} \mid \tilde{X}] \) \cite{vincent2008extracting,goodfellow2014generative}. When a block missing event overlaps an anomaly, the MSE-optimized refiner fills the gap with the most probable normal sequence and pulls the surviving fragment of the anomaly back toward normality, producing false negatives (Figure \ref{fig:imputation_bias}).

Second, on purely normal sequences, imputing large missing blocks frequently leaves unnatural reconstruction boundaries or structural artifacts, which a sensitive detector reads as abrupt deviations, producing false positives. Point Adjustment conceals this loss of precision; under strict Point-to-Point (P2P) evaluation it is directly visible as alarm fatigue for IIoT operators. We refer to these two failure modes jointly as \emph{imputation bias}, and reference this definition throughout the rest of the paper.

The objective of the data preparation phase must therefore change. Rather than aiming for statistical smoothness, a generative refiner whose sampling cost is to be paid on the edge should be aligned with the downstream detection task: it should preserve partially observed anomaly evidence and avoid fabricating artifacts on normal data. This motivates the design of the Task-Aware Diffusion Refiner (TADR).

%% file: sections/sec4.tex
\section{Proposed Methodology: TADR}
\label{sec4}

\begin{figure*}[t]
\centering
\includegraphics[width=0.75\textwidth]{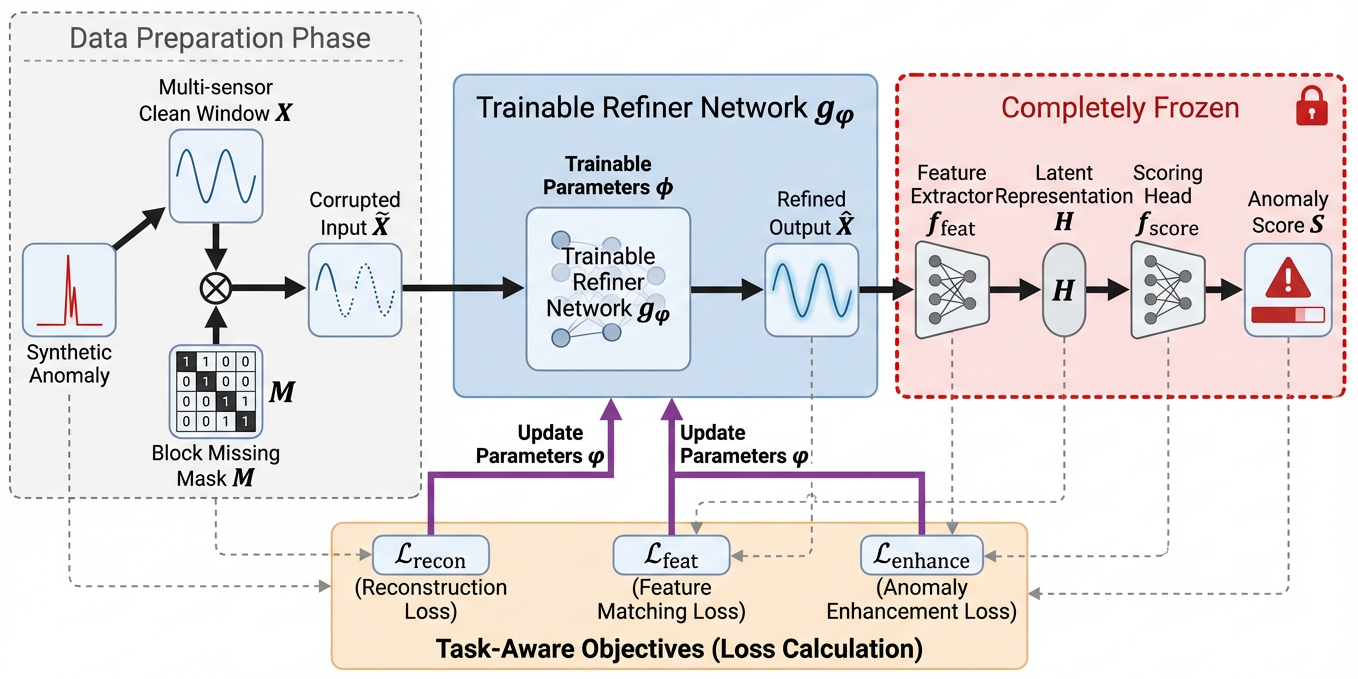}
\caption{Overall Architecture of the Task-Aware Diffusion Refiner (TADR). The corrupted input \( \tilde{X} \) is reconstructed by the trainable refiner \( g_{\phi} \). The refined output \( \hat{X} \) is then passed through the completely frozen anomaly detector to compute three task-aware objectives, ensuring that gradients flow backward solely to update the parameters \( \phi \).}
\label{fig:architecture}
\end{figure*}

To address the imputation bias and systemic inefficiencies of standard reconstruction pipelines, we propose the Task-Aware Diffusion Refiner (TADR). TADR is designed as a modular generative framework that operates directly upstream of pre-trained anomaly detectors on edge devices. Figure \ref{fig:architecture} depicts the overall architecture of the proposed methodology.

\subsection{Plug-and-Play Architecture Overview}
Let $f_{\theta}(\cdot)$ be a fully trained and fixed anomaly detector, such as Anomaly Transformer, TimesNet, or KAN-AD \cite{xu2022anomaly,wu2023timesnet,liu2024kan}. Under normal conditions, $f_{\theta}$ processes a complete window $X$ and outputs an anomaly score sequence $S \in \mathbb{R}^W$. Within our framework, the detector $f_{\theta}$ is functionally divided into two components: a feature extractor $f_{\mathrm{feat}}(\cdot)$ that generates intermediate latent representations $H \in \mathbb{R}^{W \times d}$, and a scoring head $f_{\mathrm{score}}(\cdot)$ that maps these representations to the final anomaly score $S$.

The objective is to train a generative refinement network $g_{\phi}(\cdot)$, realised as a conditional denoising diffusion model over the window. A time-conditioned 1D U-Net ($1.34$M parameters) predicts the injected noise from the concatenation of the current diffusion iterate, the observed samples masked by $M$, and the mask itself, and refinement is performed by deterministic DDIM sampling. We adopt the same quadratic noise schedule used by CSDI \cite{tashiro2021csdi} so that the comparison against that baseline isolates the training objective rather than the architecture. Given a corrupted input $\tilde{X}$, the refined output $\hat{X} = g_{\phi}(\tilde{X})$ should produce optimal latent representations and discriminative anomaly scores when processed by the frozen detector $f_{\theta}$.

Coupling a diffusion model to a downstream loss requires one non-obvious step. Back-propagating through a full $T$-step sampling trajectory is intractable, so the task-aware objectives are instead applied to the one-step estimate of the clean window implied by the current noise prediction, $\hat{X}_0 = (X_t - \sqrt{1 - \bar{\alpha}_t}\, \epsilon_{\phi}) / \sqrt{\bar{\alpha}_t}$, which the same forward pass already yields. Training therefore costs no more than plain diffusion training. Since $\hat{X}_0$ carries little signal at high noise levels, where that division amplifies any prediction error, the task-aware terms are weighted by $\bar{\alpha}_t$ and $\hat{X}_0$ is clipped to the standardised data range. Because $f_{\theta}$ remains completely frozen during the training of $g_{\phi}$, TADR maintains strict compatibility with various detectors without requiring structural modifications to the target models \cite{hinton2015distilling,romero2014fitnets}.

\subsection{Continuous-Channel Gating}
Multivariate IIoT benchmarks are not uniformly continuous: in the telemetry datasets only one channel carries an analogue sensor reading, the rest being one-hot command signals. A Gaussian diffusion process cannot represent such variables, and imputing a command is not meaningful in any case, since what a dropout destroys is the measurement. We therefore restrict the refiner to channels that are not two-level in the \emph{raw} training stream---the test must precede standardisation, which maps a binary channel to a different pair of values per entity and would make it appear continuous---and pass discrete channels through unchanged. Gating is applied identically to every refiner evaluated, so it changes the task rather than favouring a method.

\subsection{Task-Aware Noise and Anomaly Simulation}
To ensure the refiner is robust against the challenging IIoT conditions described in Section \ref{sec3}, $g_{\phi}$ is trained using a mixed-noise injection strategy \cite{devries2017improved}. During the forward pass, a clean mini-batch $X$ is augmented by a stochastic noise injector that applies block missing, random masking, or Gaussian noise based on predefined probabilities.

To train the refiner to process anomalous patterns without over-smoothing them, we introduce synthetic anomalies into a subset of the training batch. We apply a localized Gaussian perturbation (e.g., $\sigma = 5.0$) over a short, continuous span (e.g., 10 timestamps) to simulate a severe sensor malfunction or a targeted cyber-physical attack \cite{ruff2019deep}. Let $X_{\mathrm{anom}}$ denote the window injected with this synthetic anomaly, and $\tilde{X}_{\mathrm{anom}}$ denote its corresponding corrupted version.

\subsection{Task-Aware Loss Enhancer}
The primary contribution of TADR is its task-aware objective function. Rather than relying solely on the statistical distance between the input and the output, the training of $g_{\phi}$ is tightly coupled with the internal responses of the downstream detector $f_{\theta}$. The total loss function consists of three distinct components.

The first component is the reconstruction loss ($\mathcal{L}_{\mathrm{recon}}$), which ensures the basic structural restoration of the underlying normal multi-sensor dynamics:
\[ \mathcal{L}_{\mathrm{recon}} = \mathbb{E}_{X} \left[ \| X - g_{\phi}(\tilde{X}) \|_2^2 \right] \]

The second component is the feature matching loss ($\mathcal{L}_{\mathrm{feat}}$). Reconstructed data may resemble the original yet lack the channel dependencies the detector relies on, so this term forces the refiner to preserve the semantic features of the detector's intermediate layers \cite{zagoruyko2015paying,heo2019comprehensive}:
\[ \mathcal{L}_{\mathrm{feat}} = \mathbb{E}_{X} \left[ \| f_{\mathrm{feat}}(X) - f_{\mathrm{feat}}(g_{\phi}(\tilde{X})) \|_2^2 \right] \]
Since the detector parameters are fixed, gradients from this loss flow backward through $f_{\mathrm{feat}}$ solely to update the parameters of $g_{\phi}$.

The third component is the anomaly enhancement loss ($\mathcal{L}_{\mathrm{enhance}}$). This loss directly counters imputation bias by dictating distinct behaviors for normal and anomalous regions via a mathematically defined switching logic. We define an indicator variable $y \in \{0, 1\}$, where $y=1$ if the input contains a synthetic anomaly, and $y=0$ otherwise. The enhancement loss is formulated as a conditional expectation:
\[ \mathcal{L}_{\mathrm{enhance}} = \mathbb{E}_{X, y} \left[ (1 - y) \cdot \| S - \hat{S} \|_1 - y \cdot \left( \frac{1}{W} \sum_{t=1}^{W} \hat{S}_t \right) \right] \]
where $S = f_{\mathrm{score}}(X)$ acts as the reference score from the uncorrupted input, and $\hat{S} = f_{\mathrm{score}}(g_{\phi}(\tilde{X}))$ is the score from the refined input. For normal instances ($y=0$), the refiner minimizes the L1 distance strictly to prevent the generation of unnatural boundaries or artifacts that could trigger severe false positives (False Alarms). Conversely, for anomalous instances ($y=1$), the refiner is severely penalized for statistical smoothing and is driven to maximize the overall anomaly score.

The refiner $g_{\phi}$ is trained end-to-end to minimize the unified task-aware loss \cite{kendall2018multi}:
\[ \mathcal{L}_{\mathrm{total}} = \mathcal{L}_{\mathrm{recon}} + \lambda_{\mathrm{feat}} \mathcal{L}_{\mathrm{feat}} + \lambda_{\mathrm{enhance}} \mathcal{L}_{\mathrm{enhance}} \]
where $\lambda_{\mathrm{feat}}$ and $\lambda_{\mathrm{enhance}}$ are hyperparameters that balance the feature alignment and score amplification objectives. We set $\lambda_{\mathrm{feat}} = 0.1$ and $\lambda_{\mathrm{enhance}} = 0.5$; Section \ref{sec5} reports a sweep over both, which shows this setting to be statistically indistinguishable from the grid optimum and the objective to be insensitive to $\lambda_{\mathrm{feat}}$ over an order of magnitude, while $\lambda_{\mathrm{enhance}} = 0$ collapses performance.

\subsection{Training Procedure}
To ensure computational efficiency and balanced adversarial learning, the training procedure of TADR is systematically structured. As a prerequisite, the base anomaly detector $f_{\theta}$ is fully trained on the normal IIoT training dataset and its parameters are completely frozen.

During the refiner training phase, we sample mini-batches of size $B=64$ from the normal sliding windows. To stabilize the gradient updates of the conditional enhancement loss, each batch is evenly split. The first half ($B/2 = 32$) remains purely normal but is subjected to structural corruption using the mixed-noise injector. The second half is injected with the synthetic localized anomalies to create $X_{\mathrm{anom}}$, and subsequently corrupted by the same noise injector to generate $\tilde{X}_{\mathrm{anom}}$. 

Both halves are passed through the refiner $g_{\phi}$ and then evaluated by the frozen detector $f_{\theta}$. The comprehensive loss $\mathcal{L}_{\mathrm{total}}$ is computed by aggregating the normal and anomalous responses. The network is optimized using the Adam optimizer with a learning rate of $3 \times 10^{-4}$. To prevent exploding gradients caused by the adversarial nature of $\mathcal{L}_{\mathrm{enhance}}$ on the anomalous half, gradient clipping is rigorously applied with a maximum norm of 1.0. Training windows use a stride of $10$ rather than a non-overlapping stride, since a diffusion model needs far more distinct windows than a single-pass refiner: at stride $100$ a single SMD entity yields only $284$ windows and the model fails to converge, reconstructing missing regions worse than a constant fill. Across pooled datasets the epoch count is derived from the window count so every dataset receives the same gradient-step budget ($\approx 3{,}100$ updates); otherwise datasets of many short entities train an order of magnitude less than SMD and the comparison reflects budget rather than method. The refiner stays light enough for retraining on constrained edge infrastructure.

%% file: sections/sec5.tex
\section{Experimental Evaluation}
\label{sec5}

\subsection{Experimental Setup}
To validate the proposed framework, we utilize four widely used multi-sensor IIoT anomaly detection benchmarks: Server Machine Dataset (SMD), Soil Moisture Active Passive (SMAP), Mars Science Laboratory (MSL), and Pooled Server Metrics (PSM) \cite{su2019robust,hundman2018detecting,abdulaal2021practical}. To ensure a robust, global evaluation rather than cherry-picking specific subsets, we pooled all machine entities within each dataset. All continuous data sequences are processed using a sliding window size of $W=100$. Training and accuracy experiments run on a server with dual Intel Xeon E5-2620 v4 CPUs, 125 GiB of RAM, and NVIDIA GeForce RTX 3090 GPUs under Ubuntu 22.04 LTS; deployment latency is profiled separately at batch size one on a single CPU thread (Section \ref{sec:overhead}).

To simulate severe industrial operating conditions, we systematically inject continuous block missing masks, varying the missing ratio from 10\% to 50\% of the total window length. Missing ratios exceeding 30\% represent critical communication blackouts where standard statistical interpolations completely fail \cite{cao2018brits}. Additionally, we inject Gaussian noise ($\sigma=1.0$) to evaluate the framework's baseline resilience against general sensor degradation. 

Crucially, in terms of evaluation metrics, we explicitly abandon the widely criticized Point Adjustment (PA) protocol. Recent system studies indicate that PA induces a severe overestimation bias, masking the structural flaws of generative models that trigger massive false alarms across contiguous segments \cite{kim2022towards}. To accurately reflect the true system fatigue experienced in IIoT environments, we strictly report the Point-to-Point (P2P) F1-score, Precision, and Recall. All reported metrics are averaged over five independent runs with different random seeds. We report $95\%$ confidence intervals rather than means alone, and compare methods with paired $t$-tests on the per-seed differences: pairing on seed removes the variance the two methods share, which is the comparison the claims actually rest on. Because a favourable best-threshold F1 does not by itself establish that a refiner reconstructed anything, we additionally record reconstruction fidelity---output scale, the fraction of samples pinned to the clipping bound, and reconstruction error on observed and missing positions---for every configuration, and report it alongside detection quality.

\subsection{Mitigation of Imputation Bias (Precision-Recall Imbalance)}
\begin{figure}[ht]
\centering
\includegraphics[width=0.98\columnwidth]{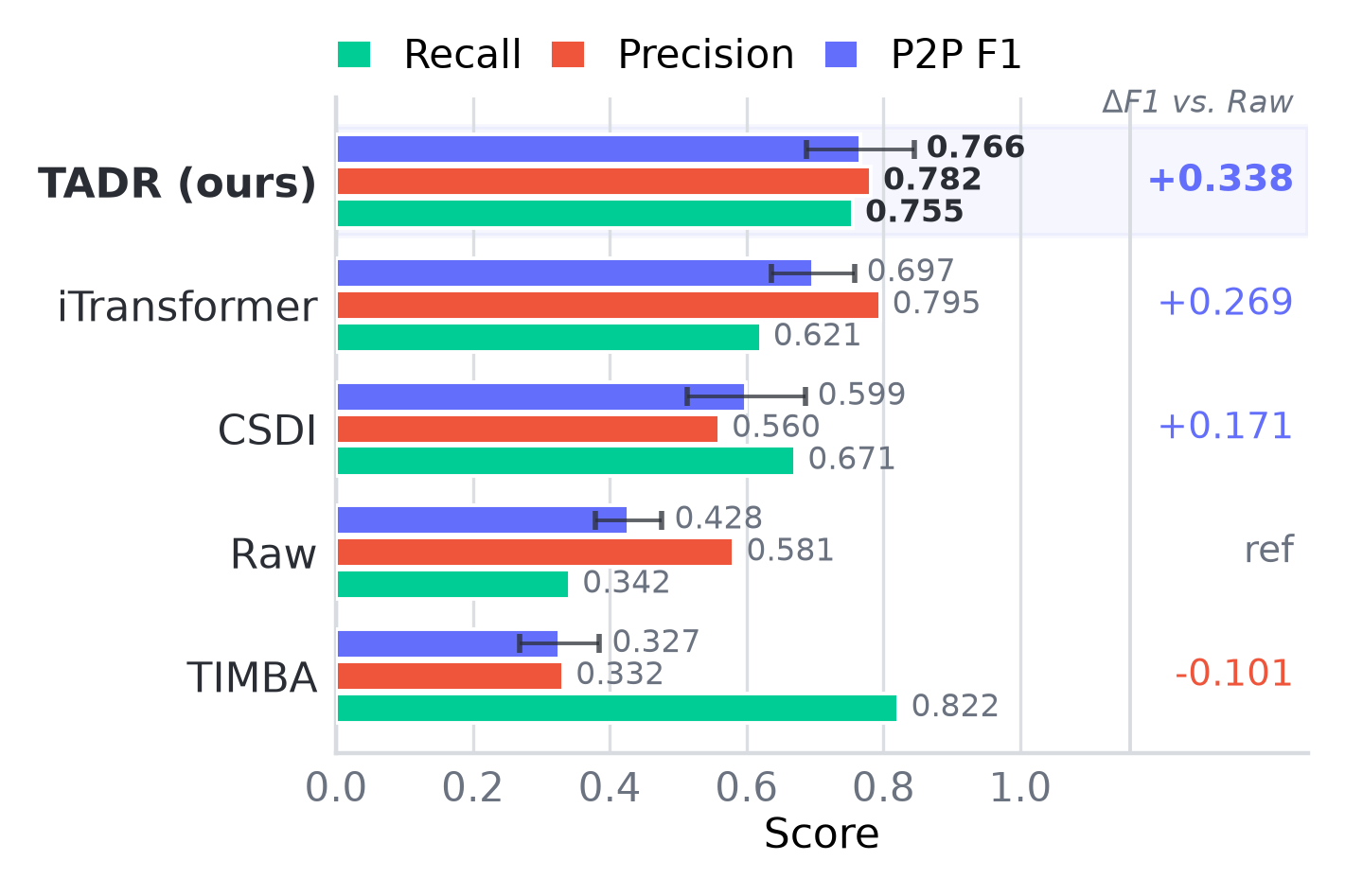}
\caption{Refiner comparison under 40\% block missing (SMD 1-1). Means over five seeds; error bars are 95\% intervals on F1, and the right column the paired difference against the unrefined detector (paired $t$-test: TADR $p = 0.0015$, iTransformer $p = 0.0010$, CSDI $p = 0.018$, TIMBA $p = 0.028$).}
\label{fig:refiner_comparison}
\end{figure}

To isolate the effect of pure error-centric imputation, we evaluate the baselines under a severe 40\% block missing scenario. Figure \ref{fig:refiner_comparison} reports the comparison on a representative subset (SMD 1-1), where exact model behaviours are visible.

\begin{figure*}[ht]
\centering
\includegraphics[width=0.8\textwidth]{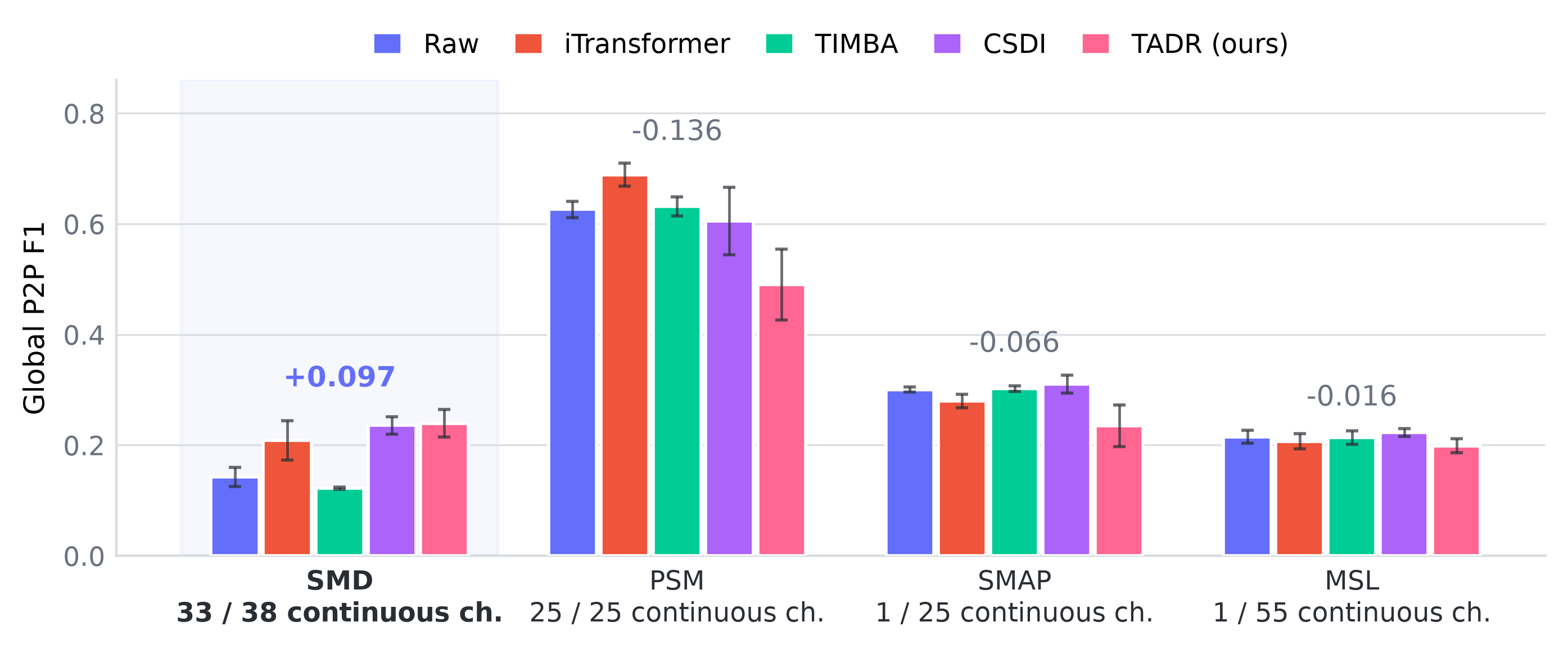} 
\caption{Global P2P F1 across four IIoT datasets at 40\% block missing, five seeds with 95\% confidence intervals. Datasets are annotated with the number of continuous channels they provide, which is what the benefit of task-aware refinement tracks.}
\label{fig:global_f1_40}
\end{figure*}

Imputation bias is visible most clearly in TIMBA, which attains $0.822$ recall at $0.332$ precision and ends up \emph{below} the unrefined detector ($p = 0.028$): MSE-optimised reconstruction fills missing blocks with smooth normal patterns, creating abrupt boundary artifacts that the downstream detector reads as widespread anomalies. Under the relaxed PA metric such a model would appear effective; strict P2P evaluation exposes the precision cost that drives alarm fatigue.

This failure is not universal among reconstruction-based refiners: CSDI and iTransformer both improve substantially over the unrefined detector. The reading these numbers support is not that MSE-trained imputation fails outright, but that aligning the refinement objective with the detector yields a further gain on top of it. TADR reaches the highest F1 ($+0.338$ against the unrefined detector, $p = 0.0015$) and the highest precision simultaneously, indicating the additional signal it recovers is not bought with false alarms.

\subsection{Where the Recovered Detections Come From}
A refiner can only add detections where some anomaly evidence survives the outage. To test whether TADR recovers partially observed anomalies or infers unseen ones from context, we bucket every anomalous point of the SMD 1-1 comparison by its \emph{masked fraction}: the share of the sliding windows covering that point in which it was dropped. Table \ref{tab:coverage} reports the detection rate per bucket at each method's F1-optimal threshold.

\begin{table}[ht]
\centering
\caption{Detection rate by masked fraction, the share of covering windows in which an anomalous point was dropped (SMD 1-1, 40\% block missing, five seeds). The last row is the false-alarm rate on normal points.}
\label{tab:coverage}
\begin{tabular}{l r c c c}
\hline
Masked fraction & Points & Raw & CSDI & TADR \\
\hline
25--50\% & 2,643 & 0.344 & 0.735 & \textbf{0.778} \\
50--75\% & 51 & 0.205 & 0.665 & \textbf{0.830} \\
75--100\% & 0 & --- & --- & --- \\
\hline
Normal (false alarms) & 25,785 & 0.027 & 0.056 & 0.037 \\
\hline
\end{tabular}
\end{table}

No anomalous point was masked in more than $75\%$ of its windows: each window draws its own mask, so under stride-one evaluation every point is observed in many windows. A point that no window ever observes is unrecoverable in principle, and our results make no claim about that case. Within the observed range, TADR's advantage over the unrefined detector grows with the masked fraction ($+0.43$ to $+0.63$), the opposite of what an account based on inferring the anomaly from its context would predict, although the higher bucket holds only 51 points. Its false-alarm rate stays close to the unrefined detector's and below CSDI's, so the additional detections are not bought by inflating scores. The mechanism is therefore the preservation of partially observed anomaly evidence, and the method assumes that each anomaly remains observed in at least part of the windows that cover it.

\subsection{Generality Across Detector Architectures}
We further investigate the architectural generality of the task-aware mechanism. We integrated TADR directly upstream of three distinct state-of-the-art detector architectures (Anomaly Transformer, TimesNet, and KAN-AD). 

\begin{figure}[ht]
\centering
\includegraphics[width=\columnwidth]{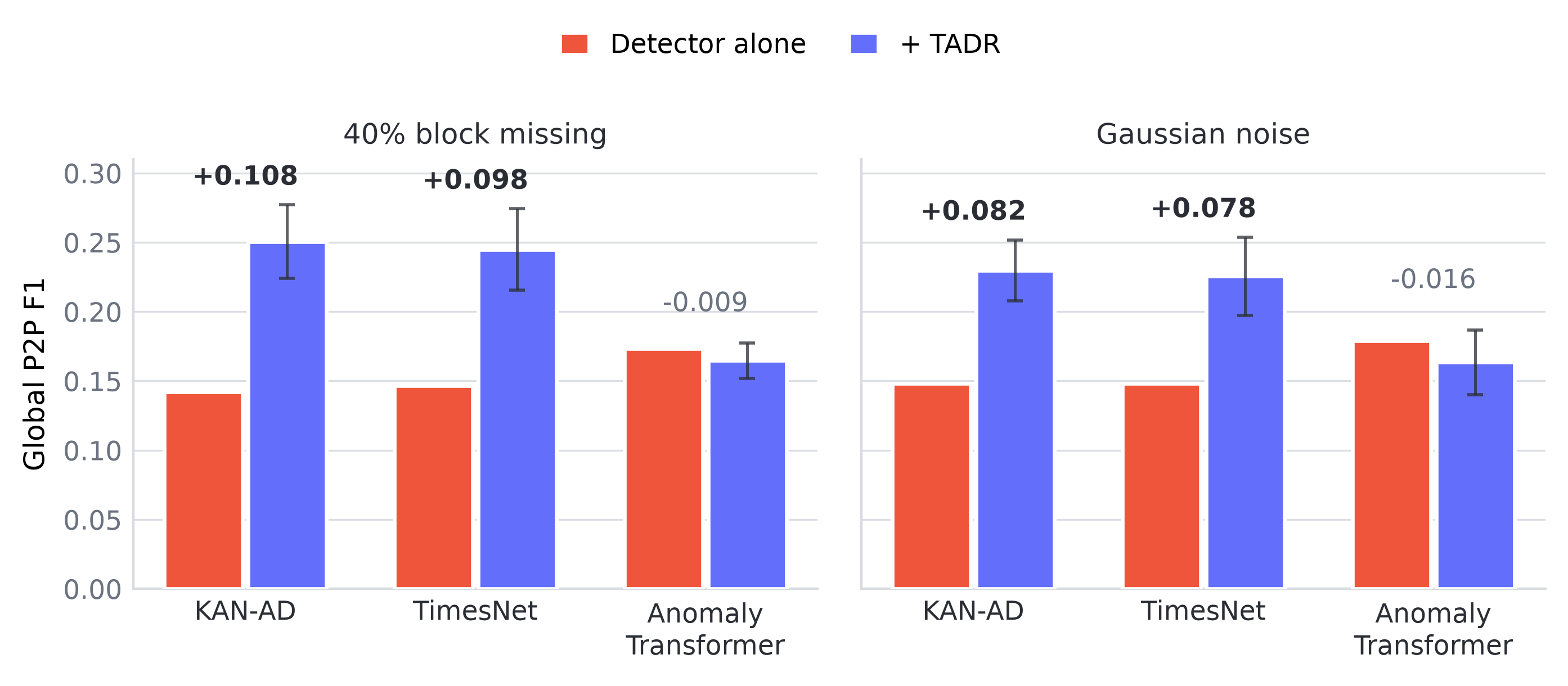}
\caption{Detector generality on pooled SMD. Bars are means over five seeds, error bars 95\% intervals on the refined score, and the annotation above each pair the paired difference. Gains are significant for KAN-AD and TimesNet ($p \le 0.002$), absent for Anomaly Transformer.}
\label{fig:plugin_robustness}
\end{figure}

Figure \ref{fig:plugin_robustness} reports the result, and it is more selective than we had expected. On the two reconstruction-error detectors the gain is large and significant: $+0.108$ for KAN-AD and $+0.098$ for TimesNet under 40\% block missing ($p = 0.0003$, $p = 0.0007$), and $+0.082$ and $+0.078$ under Gaussian noise. On Anomaly Transformer it is absent in both conditions ($-0.009$, $-0.016$; neither significant).

We therefore describe TADR as a plug-in for reconstruction-error detectors rather than a universally applicable optimizer. Both architectures that benefit score a window by how badly they reconstruct it, so sharpening a suppressed anomaly moves the score directly, whereas Anomaly Transformer derives its score from association discrepancy, which a refined window need not shift in the same way---consistent with reports that detectors respond to noise in structurally different ways \cite{park2025robustness}. The Gaussian condition doubles as a sanity check: with no missing block to fill, the refiner neither helps nor harms, confirming it does not degrade a pipeline already on intact data.

\subsection{What the Gain Depends On}
\label{sec:depends}
Task-aware refinement reconstructs a corrupted window from the channels that survived, so its benefit should scale with how much mutually redundant continuous signal remains. We test that prediction directly, by withholding channels from a dataset where the method works.

\begin{table}[ht]
\centering
\caption{Benefit against the number of channels retained (SMD 1-1, 40\% block missing), and against the overlap between successive evaluation windows. $\Delta$ is the paired difference over five seeds.}
\label{tab:redundancy}
\resizebox{\columnwidth}{!}{%
\begin{tabular}{l c c c c c c}
\hline
Channels retained & 38 & 24 & 12 & 6 & 3 & 1 \\
$\Delta$ F1 vs.\ Raw & \textbf{+0.356} & +0.293 & +0.217 & +0.168 & +0.050 & +0.041 \\
$p$ & $<$0.0001 & 0.004 & 0.015 & 0.049 & 0.51 & 0.29 \\
\hline
Test stride & 1 & 5 & 25 & 50 & 100 & \\
$\Delta$ F1 vs.\ Raw & +0.352 & +0.372 & +0.380 & +0.339 & \textbf{+0.300} & \\
$p$ & $<$0.0001 & $<$0.0001 & $<$0.0001 & $<$0.0001 & 0.0004 & \\
\hline
\end{tabular}%
}
\end{table}

The two forms of redundancy behave differently. Spatial redundancy carries the method: the benefit falls monotonically as channels are withheld, from $+0.356$ with all $38$ to $+0.041$ with one, and is no longer significant below six channels. Temporal overlap does not: scoring each timestamp exactly once, at a test stride equal to the window length, leaves the gain essentially intact, so the overlapped sliding window is a convenience of the protocol rather than part of the mechanism.

This predicts that the gain tracks the continuous channels a dataset actually provides, not its nominal count. We pool all entities within each of the four datasets and test that as the missing ratio rises from 10\% to 50\%.

The prediction holds (Figure \ref{fig:global_f1_40}). SMD supplies $33$ of $38$ continuous channels and is where TADR is strongest, with a paired gain of $+0.090$ to $+0.108$, significant at every missing ratio from 10\% to 50\% ($p \le 0.0014$) and flat as the ratio rises. At this pooled scale a plain diffusion imputer is nearly as effective ($+0.087$ to $+0.093$), so the margin of the task-aware objective over CSDI is clearest on the single entity, where it is $0.17$ (Figure \ref{fig:refiner_comparison}), and in the channel ablation, where it is $0.12$ with all channels retained. The telemetry datasets are the opposite of what their channel counts suggest: SMAP and MSL nominally offer $25$ and $55$ channels, but only one of each is an analogue sensor reading, the rest being one-hot command signals a Gaussian process cannot represent and a dropout does not meaningfully destroy. With one continuous stream to work from they sit at the far-right end of the channel row of Table \ref{tab:redundancy}, exactly where the controlled ablation places them. Continuous-channel gating keeps the refiner off the command channels but cannot manufacture redundancy the data lacks: at 40\% missing TADR remains below the unrefined detector on SMAP ($-0.066$) and MSL ($-0.016$).

PSM sharpens the condition further, since its channels are entirely continuous yet the gain is also absent. There the unrefined detector already attains $0.88$ recall, so little anomaly evidence is lost to the outage in the first place; a plain diffusion imputer is roughly neutral on the same data, which locates the cost in the task-aware objective rather than in generative refinement. Redundancy is therefore necessary but not sufficient: the benefit scales with how much evidence the outage actually removes from the detector, and vanishes when little is removed.

\subsection{Ablation Study on Task-Aware Objectives}
To rigorously validate the contribution of each objective within the TADR framework, we conducted an ablation study under the 40\% block missing scenario. To ensure a strict yet stable evaluation of exact timestamps, we utilized a smoothed P2P metric (moving average window of 5). 

\begin{table}[ht]
\centering
\caption{Ablation on loss components (smoothed P2P F1, 40\% block missing). $\Delta$ is the paired difference against the MSE-only refiner over five seeds.}
\label{tab:ablation}
\resizebox{\columnwidth}{!}{%
\begin{tabular}{l c c c c c}
\hline
Loss Components & P2P F1 & Precision & Recall & $\Delta$ & $p$ \\
\hline
Recon (MSE Only) & $0.306 \pm 0.069$ & 0.525 & 0.313 & --- & --- \\
Recon + Feature & $0.542 \pm 0.091$ & 0.561 & 0.540 & $+0.236$ & 0.012 \\
\textbf{Recon + Feature + Enhance (Full)} & $\mathbf{0.769 \pm 0.049}$ & \textbf{0.762} & \textbf{0.787} & $\mathbf{+0.463}$ & 0.0002 \\
\hline
Full objective, single-pass refiner & $0.469 \pm 0.065$ & 0.708 & 0.359 & $+0.163$ & 0.010 \\
\hline
\end{tabular}%
}
\end{table}

Table \ref{tab:ablation} shows a clear progression. The MSE-only refiner yields $0.306$, held back by imputation bias. Feature matching contributes $+0.236$ by aligning the latent representations, but it buys recall at the cost of precision ($0.525 \to 0.561$ against $0.313 \to 0.540$ recall), because boundary artifacts still trigger false alarms. The decisive component is the anomaly enhancement loss, which raises F1 to $0.769$---a paired gain of $+0.463$ over the MSE-only refiner ($p = 0.0002$), the largest effect measured anywhere in this work---and, unlike the intermediate step, improves precision and recall together. The final row isolates the generative process itself: a deterministic single-pass refiner trained on the identical objective and data reaches only $0.469$, so the iterative formulation is worth $+0.300$ beyond the loss function, which is what justifies its sampling cost.

\subsection{Sensitivity to the Task-Aware Loss Weights}
\begin{figure}[ht]
\centering
\includegraphics[width=\columnwidth]{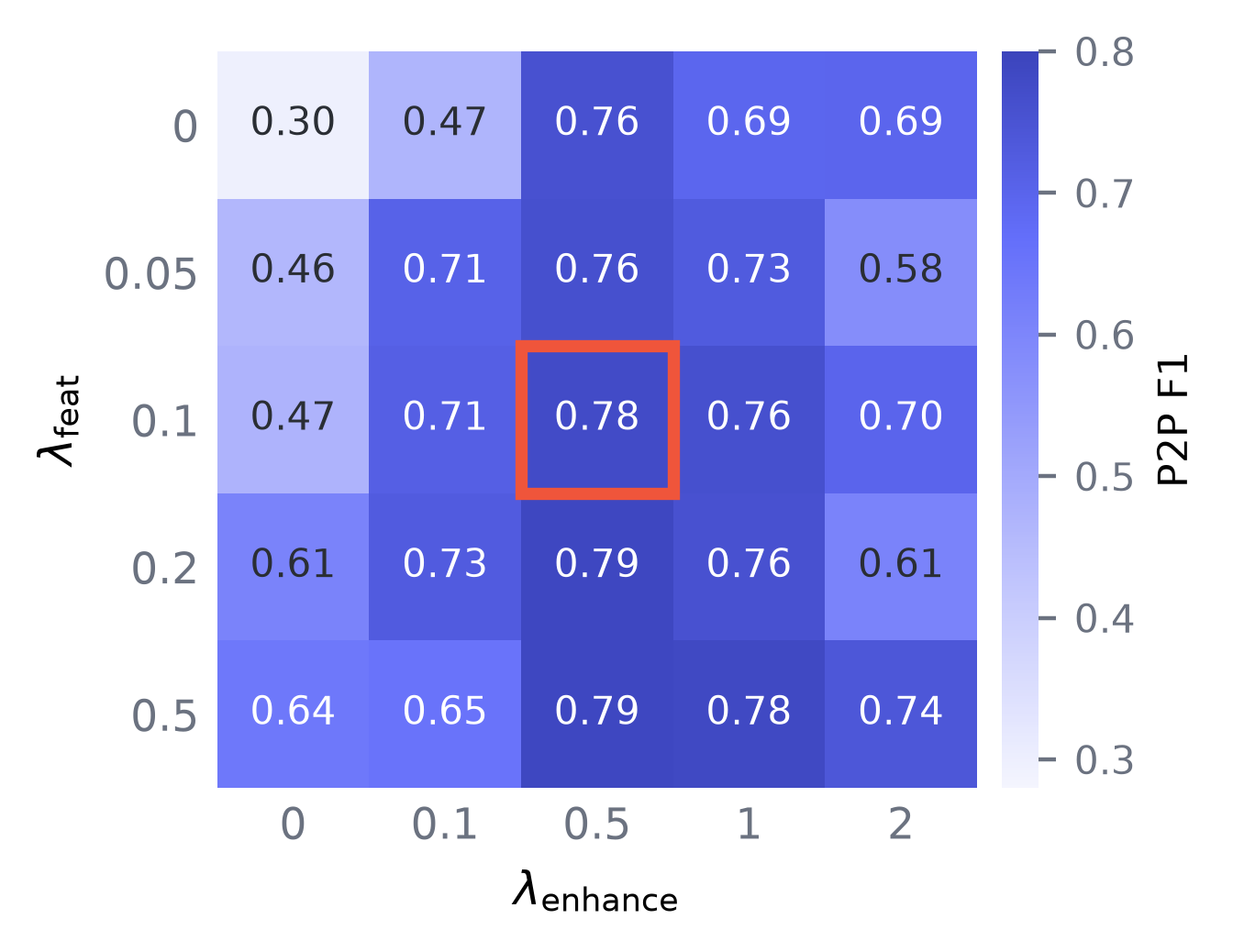}
\caption{P2P F1 over the $(\lambda_{\mathrm{feat}}, \lambda_{\mathrm{enhance}})$ grid, mean of five seeds. The setting used throughout the paper is outlined.}
\label{fig:lambda}
\end{figure}

The two weights $\lambda_{\mathrm{feat}}$ and $\lambda_{\mathrm{enhance}}$ are the only hyperparameters the task-aware objective introduces, and Figure \ref{fig:lambda} sweeps both over a $5 \times 5$ grid under the same protocol as the ablation.

The setting used throughout, $(0.1, 0.5)$, attains $0.776$ against a grid optimum of $0.787$---a difference well inside the confidence interval---and the objective is insensitive to $\lambda_{\mathrm{feat}}$ over an order of magnitude at the chosen $\lambda_{\mathrm{enhance}}$. Most informative is the leftmost column: with $\lambda_{\mathrm{enhance}} = 0$ the grid collapses to between $0.296$ and $0.639$ regardless of $\lambda_{\mathrm{feat}}$, independently confirming that the enhancement term, not feature matching, carries the method. Beyond $\lambda_{\mathrm{enhance}} = 1.0$ performance degrades again as the refiner begins inflating scores on normal windows.

\subsection{System Overhead and Edge Feasibility}
\label{sec:overhead}
Diffusion makes inference cost proportional to the sampling budget, scaling as $\mathcal{O}(T \cdot \mathcal{C}_{\mathrm{CNN}})$. We therefore report latency against $T$, and per window as well as per batch, since the two differ by two orders of magnitude and only the former is meaningful for streaming.

\begin{table}[ht]
\centering
\caption{Inference overhead. Batch figures use batch size 256 on one RTX 3090 with CUDA synchronisation, measured with the GPU otherwise idle (means of five runs; the $T{=}10$ row is from the sampling-step profile). Streaming rows use batch size 1 on a single CPU thread of a 2.1 GHz Xeon E5-2620 v4. The refiner occupies 4.03 MB.}
\label{tab:system_overhead}
\resizebox{\columnwidth}{!}{%
\begin{tabular}{l c c}
\hline
Configuration & ms / batch & ms / window \\
\hline
Raw (No Refiner) & 0.39 & 0.002 \\
TIMBA & 0.85 & 0.003 \\
iTransformer & 1.67 & 0.007 \\
CSDI ($T{=}50$) & 241.8 & 0.944 \\
Ours (TADR), $T{=}50$ & 240.5 & 0.939 \\
\textbf{Ours (TADR), $T{=}10$} & \textbf{44.2} & \textbf{0.173} \\
\hline
\multicolumn{3}{l}{\textit{Streaming, one CPU core (1 Hz sensor streams sustained)}} \\
TADR $T{=}50$ / $T{=}10$ / $T{=}5$ & --- & 205.6 / \textbf{41.4} / 20.8 \\
\quad concurrent streams per core & --- & 5 / \textbf{24} / 48 \\
\hline
\end{tabular}%
}
\end{table}

Non-generative refiners execute in about a millisecond per batch while diffusion pays for iterative sampling: roughly $240$ ms at the full budget, the same as CSDI, with which TADR shares its network and sampler. Ten steps brings this to $44.2$ ms per batch, and $41.4$ ms per window in streaming mode, at which point a single CPU core sustains roughly $24$ concurrent $1$ Hz streams from a $4.03$ MB model---so a GPU is not a prerequisite. The single server core is a proxy for a gateway-class processor; we have not measured on an embedded board, where per-core throughput is typically lower.

\subsection{Sampling Budget and Reconstruction Fidelity}
Because latency is linear in $T$, the sampling budget is the principal deployment lever. Table \ref{tab:step_budget} sweeps it with reconstruction fidelity reported next to detection quality.

\begin{table}[ht]
\centering
\caption{Detection quality and reconstruction fidelity against the sampling budget (SMD 1-1, 40\% block missing). Saturation is the fraction of output pinned to the clipping bound; observed MSE is over never-missing samples.}
\label{tab:step_budget}
\begin{tabular}{c c c c c}
\hline
$T$ & CSDI F1 & TADR F1 & TADR sat.\ (\%) & TADR obs.\ MSE \\
\hline
50 & 0.628 & \textbf{0.710} & 0.0 & 1.86 \\
25 & 0.564 & \textbf{0.714} & 0.0 & 1.64 \\
10 & 0.426 & \textbf{0.747} & 0.0 & 1.96 \\
5 & 0.452 & 0.710 & 0.0 & 6.16 \\
2 & 0.435 & 0.695 & 0.9 & 63.0 \\
\hline
\end{tabular}
\end{table}

TADR is statistically flat between ten and fifty steps: the paired difference is $+0.038$ with $p = 0.60$, indistinguishable at five seeds. CSDI degrades sharply below twenty-five ($0.426$ at ten against $0.628$ at fifty). We attribute this to where the task-aware terms act---on the one-step estimate $\hat{X}_0$, exactly what few-step sampling relies on, so the model is trained to make that estimate useful to the detector. Since latency is linear in $T$, $T=10$ is the operating point we recommend: a five-fold saving at no measurable cost in quality.

The lower rows matter for a different reason. At two steps the output saturates its clipping bound on roughly half of all samples and reconstruction error rises fortyfold, yet best-threshold P2P F1 remains near $0.69$. A detection metric that cannot separate this from genuine refinement is not sufficient evidence alone---the same weakness we identify in Point Adjustment, and the reason fidelity is reported throughout.

%% file: sections/sec6.tex
\section{Discussion}
\label{sec6}

\subsection{Edge Deployment and System Acceleration}
TADR inherits the iterative sampling cost of its generative architecture: $44$ ms per batch of $256$ windows at ten steps, rising to about $240$ ms at fifty. The question that matters for deployment is not throughput but per-stream cost on the device that would run it, and a single CPU core sustains roughly $24$ concurrent $1$ Hz streams at ten steps with a $4.03$ MB model (Section \ref{sec5}). GPU acceleration is therefore not a prerequisite, though power, thermal headroom, and multi-tenant scheduling on PLCs and low-power gateways remain open \cite{croce2020reliable}.

Two further directions remain, and our measurements qualify both. Reducing the sampling budget is the more effective lever, since latency is linear in $T$, and it realises much of what step distillation or Consistency Models \cite{song2023consistency} promise without additional training---but it has a floor. One-step sampling from pure noise must divide by $\sqrt{\bar{\alpha}_T} \approx 0.006$, and the output then saturates its clipping bound on over half of all samples while still scoring a competitive best-threshold F1, which is why fidelity accompanies every detection metric. Compiler-level quantisation is the second, but must be the right kind: \emph{dynamic} INT8 is marginally \emph{slower} than FP32 here ($47.3$ against $41.4$ ms), targeting linear layers while this refiner's cost is convolutional. Static or quantisation-aware training is the appropriate route. 

Because quality is flat between fifty and ten steps but collapses below five, $T$ is also a natural control variable, lowered under power or thermal pressure and raised when headroom returns, above the degeneracy floor we identify.

\subsection{Structural Redundancy and Graceful Degradation}
On SMD, TADR's gain is flat across block missing ratios from 10\% to 50\% (Section \ref{sec:depends}), an instance of graceful degradation obtained by exploiting structural redundancy. Section \ref{sec5} attributes it to spatial rather than temporal redundancy, and the practical consequence is a design one: because the benefit depends on how many mutually redundant channels a deployment carries rather than on how densely it samples, a topology with overlapping receptive fields tolerates outages that would otherwise demand duplicated hardware.

\subsection{Limitations and Future Work}
While TADR proves highly effective for continuous, physically correlated multi-sensor streams, its current architecture assumes the presence of several mutually redundant continuous channels, and we observed two regimes in which it does not help.

The first is data dominated by discrete channels. A Gaussian diffusion process has no way to represent one-hot command signals, and restricting the refiner to the continuous channels recovers most, but not all, of the deficit on the telemetry datasets. The same limitation applies a fortiori to asynchronous event logs such as network routing or financial transaction streams.

The second is more interesting and less expected. On PSM, whose channels are entirely continuous, the unrefined detector already attains $0.88$ recall, so there is little missed signal for the refiner to restore and refinement subtracts from what the detector would otherwise see. We attempted nine variants targeting this regime---rescaling, budget, both loss weights, the synthetic anomaly length and magnitude, its batch fraction, and confining the refiner to the missing region---and none recovered it. A plain diffusion imputer without the task-aware terms is neutral on this dataset, which locates the cost in the task-aware machinery rather than in generative refinement as such. We therefore state the operating condition directly: the benefit scales with how much anomaly evidence the detector loses to the outage, and vanishes when little is lost.

Two threats remain outside our evaluation. TADR is resilient to random Gaussian noise, but a targeted poisoning or adversarial perturbation injected into the observable channels could steer the refiner into hallucinating boundary artifacts and bypassing the detector; bounding the confidence of generated imputations, for instance via conformal prediction, is the natural response. Industrial systems also undergo concept drift through mechanical wear, and adapting the task-aware alignment over time without full retraining remains open. Future work will address asynchronous and heterogeneous sensor networks, zero-shot drift adaptation, and multi-tenant edge scheduling when several detection pipelines share one accelerator under QoS constraints.

%% file: sections/sec7.tex
\section{Conclusion}
\label{sec7}

In this study, we addressed the systemic inefficiencies and structural vulnerabilities of conventional time-series anomaly detection pipelines deployed in IIoT edge environments subject to severe block missing conditions. We showed that the standard ``impute-then-detect'' paradigm, which relies on disjointed MSE-optimized generative models, admits an imputation bias that smooths over cyber-physical attack signatures and generates boundary artifacts that trigger alarm fatigue.

To resolve this, we introduced TADR, a conditional diffusion refiner whose training objective is coupled to the loss landscape of the frozen downstream detector. Because back-propagating through the sampling trajectory is intractable, the task-aware terms are applied to the one-step estimate of the clean window that the noise prediction already provides, leaving training no more expensive than plain diffusion training. Under strict Point-to-Point evaluation with five seeds and paired significance tests, TADR raises P2P F1 by $+0.338$ over the unrefined detector on SMD, and by $+0.090$ to $+0.108$ on pooled SMD at every missing ratio from 10\% to 50\%. The ablation isolates the source of this gain: the anomaly enhancement objective alone contributes $+0.463$, and the iterative generative formulation contributes a further $+0.300$ over a single-pass refiner trained on the identical objective.

Equally important is where the method does not help. The benefit scales with how much anomaly evidence the outage removes from the detector: it is large on multivariate continuous streams with mutually redundant channels, and disappears when the observable channels are predominantly one-hot command signals or when the unrefined detector already recovers most anomalies unaided. We report these regimes explicitly, with a controlled channel ablation reproducing the same collapse on a dataset where the method does work, because a precisely stated operating condition is more useful than an averaged claim of superiority.

On feasibility, a GPU is not a prerequisite: at ten sampling steps the refiner retains its detection quality while running in $41.4$ ms per window on a single CPU core, sustaining roughly $24$ concurrent 1 Hz streams from a $4.03$ MB model. We also identify the budget below which sampling degenerates while conventional detection metrics fail to register it, which is why reconstruction fidelity is reported throughout.

This work argues against separating reconstruction from detection in resource-constrained pipelines, and for stating the conditions under which coupling them pays. Future work will address discrete and asynchronous channels, adaptive sampling under power and thermal limits, and validation on physical testbeds under operational drift.

%% file: references/references.bib
@inproceedings{hundman2018detecting,
  title={Detecting spacecraft anomalies using lstms and nonparametric dynamic thresholding},
  author={Hundman, Kyle and Constantinou, Valentino and Laporte, Christopher and Colwell, Ian and Soderstrom, Tom},
  booktitle={Proceedings of the 24th ACM SIGKDD international conference on knowledge discovery \& data mining},
  pages={387--395},
  year={2018}
}

@inproceedings{miao2021generative,
  title={Generative semi-supervised learning for multivariate time series imputation},
  author={Miao, Xiaoye and Wu, Yangyang and Wang, Jun and Gao, Yunjun and Mao, Xudong and Yin, Jianwei},
  booktitle={Proceedings of the AAAI Conference on Artificial Intelligence},
  volume={35},
  pages={8983--8991},
  year={2021}
}

@article{luo2018multivariate,
  title={Multivariate time series imputation with generative adversarial networks},
  author={Luo, Yonghong and Cai, Xiangrui and Zhang, Ying and Xu, Jun and Yuan, Xiaojie},
  journal={Advances in neural information processing systems},
  volume={31},
  year={2018}
}

@article{tashiro2021csdi,
  title={Csdi: Conditional score-based diffusion models for probabilistic time series imputation},
  author={Tashiro, Yusuke and Song, Jiaming and Song, Yang and Ermon, Stefano},
  journal={Advances in Neural Information Processing Systems},
  volume={34},
  pages={24804--24816},
  year={2021}
}

@inproceedings{xu2022anomaly,
  title={Anomaly transformer: Time series anomaly detection with association discrepancy},
  author={Xu, Jiehui and Wu, Haixu and Wang, Jianmin and Long, Mingsheng},
  booktitle={International Conference on Learning Representations (ICLR)},
  year={2022}
}

@inproceedings{wu2023timesnet,
  title={TimesNet: Temporal 2D-variation modeling for general time series analysis},
  author={Wu, Haixu and Hu, Tengge and Liu, Yong and Dong, Hang and Wang, Jianmin and Long, Mingsheng},
  booktitle={International Conference on Learning Representations (ICLR)},
  year={2023}
}

@inproceedings{zong2018deep,
  title={Deep autoencoding Gaussian mixture model for unsupervised anomaly detection},
  author={Zong, Bo and Song, Qi and Min, Martin Renqiang and Cheng, Wei and Lumezanu, Cristian and Cho, Daeki and Chen, Haifeng},
  booktitle={International Conference on Learning Representations (ICLR)},
  year={2018}
}

@inproceedings{audibert2020usad,
  title={Usad: Unsupervised anomaly detection on multivariate time series},
  author={Audibert, Julien and Michiardi, Pietro and Guyard, Fr{\'e}d{\'e}ric and Marti, S{\'e}bastien and Zuluaga, Maria A},
  booktitle={Proceedings of the 26th ACM SIGKDD international conference on knowledge discovery \& data mining},
  pages={3395--3404},
  year={2020}
}

@inproceedings{li2019mad,
  title={MAD-GAN: Multivariate anomaly detection for time series data with generative adversarial networks},
  author={Li, Dan and Chen, Dacheng and Jin, Baihong and Shi, Lei and Goh, Jonathan and Ng, See-Kiong},
  booktitle={International conference on artificial neural networks},
  pages={703--716},
  year={2019},
  organization={Springer}
}

@inproceedings{rasul2021autoregressive,
  title={Autoregressive denoising diffusion models for multivariate probabilistic time series forecasting},
  author={Rasul, Kashif and Seward, Calvin and Schuster, Ingmar and Vollgraf, Roland},
  booktitle={International Conference on Machine Learning},
  pages={8857--8868},
  year={2021},
  organization={PMLR}
}

@inproceedings{su2019robust,
  title={Robust anomaly detection for multivariate time series through stochastic recurrent neural network},
  author={Su, Ya and Zhao, Youjian and Niu, Chenhao and Liu, Rong and Sun, Wei and Pei, Dan},
  booktitle={Proceedings of the 25th ACM SIGKDD international conference on knowledge discovery \& data mining},
  pages={2828--2837},
  year={2019}
}

@inproceedings{abdulaal2021practical,
  title={Practical approach to asynchronous multivariate time series anomaly detection and localization},
  author={Abdulaal, Ahmed and Liu, Zhuanghua and Lance, Tom},
  booktitle={Proceedings of the 27th ACM SIGKDD conference on knowledge discovery \& data mining},
  pages={2465--2474},
  year={2021}
}

@inproceedings{malhotra2015long,
  title={Long short term memory networks for anomaly detection in time series},
  author={Malhotra, Pankaj and Vig, Lovekesh and Shroff, Gautam and Agarwal, Puneet},
  booktitle={Proceedings of the 23rd European symposium on artificial neural networks, computational intelligence and machine learning (ESANN)},
  pages={89},
  year={2015}
}

@inproceedings{park2018multimodal,
  title={A multimodal anomaly detector for robot-assisted feeding using an lstm-based variational autoencoder},
  author={Park, Daehyung and Hoshi, Yuuna and Kemp, Charles C},
  booktitle={IEEE Robotics and Automation Letters},
  volume={3},
  pages={1544--1551},
  year={2018},
  publisher={IEEE}
}

@inproceedings{liu2024itransformer,
  title={iTransformer: Inverted Transformers Are Effective for Time Series Forecasting},
  author={Liu, Yong and Hu, Tengge and Zhang, Haoran and Wu, Haixu and Wang, Shiyu and Ma, Lintao and Wang, Jianmin},
  booktitle={International Conference on Learning Representations (ICLR)},
  year={2024}
}

@inproceedings{liu2024kan,
  title={KAN: Kolmogorov-Arnold Networks},
  author={Liu, Ziming and Wang, Yixuan and Vaidya, Sachin and Ruehle, Fabian and Halverson, James and Solja{\v{c}}i{\'c}, Marin and Hou, Thomas Y and Tegmark, Max},
  booktitle={International Conference on Machine Learning (ICML)},
  year={2024}
}

@inproceedings{tang2020joint,
  title={Joint modeling of local and global temporal dynamics for multivariate time series anomaly detection with partial observations},
  author={Tang, Yaqi and Zhao, Yifan and Liu, Yu and Pei, Dan},
  booktitle={Proceedings of the AAAI Conference on Artificial Intelligence},
  volume={34},
  pages={5956--5963},
  year={2020}
}

@book{little2019statistical,
  title={Statistical analysis with missing data},
  author={Little, Roderick JA and Rubin, Donald B},
  volume={793},
  year={2019},
  publisher={John Wiley \& Sons}
}

@article{che2018recurrent,
  title={Recurrent neural networks for multivariate time series with missing values},
  author={Che, Zhengping and Purushotham, Sanjay and Cho, Kyunghyun and Sontag, David and Liu, Yan},
  journal={Scientific reports},
  volume={8},
  number={1},
  pages={1--12},
  year={2018},
  publisher={Nature Publishing Group}
}

@inproceedings{yoon2018estimate,
  title={Time-series generative adversarial networks},
  author={Yoon, Jinsung and Jarrett, Daniel and Van der Schaar, Mihaela},
  booktitle={Advances in Neural Information Processing Systems (NeurIPS)},
  volume={32},
  year={2019}
}

@inproceedings{cao2018brits,
  title={BRITS: Bidirectional Recurrent Imputation for Time Series},
  author={Cao, Wei and Wang, Dong and Li, Jian and Zhou, Hao and Li, Lei and Li, Yanyan},
  booktitle={Advances in Neural Information Processing Systems (NeurIPS)},
  volume={31},
  year={2018}
}

@article{du2023saits,
  title={SAITS: Self-attention-based imputation for time series},
  author={Du, Wenjie and Cote, David and Liu, Yan},
  journal={Expert Systems with Applications},
  volume={219},
  pages={119619},
  year={2023},
  publisher={Elsevier}
}

@inproceedings{zhao2020multivariate,
  title={Multivariate time-series anomaly detection via graph attention network},
  author={Zhao, Hang and Wang, Yujing and Duan, Ju and Huang, Congrui and Cao, Defu and Tong, Yunhai and Xu, Bixiong and Bai, Jingbai and Tong, Jie and Zhang, Qi},
  booktitle={International Conference on Data Mining (ICDM)},
  pages={841--850},
  year={2020},
  organization={IEEE}
}

@article{ho2020denoising,
  title={Denoising diffusion probabilistic models},
  author={Ho, Jonathan and Jain, Ajay and Abbeel, Pieter},
  journal={Advances in neural information processing systems},
  volume={33},
  pages={6840--6851},
  year={2020}
}

@inproceedings{song2020score,
  title={Score-based generative modeling through stochastic differential equations},
  author={Song, Yang and Sohl-Dickstein, Jascha and Kingma, Diederik P and Kumar, Abhishek and Ermon, Stefano and Poole, Ben},
  booktitle={International Conference on Learning Representations (ICLR)},
  year={2021}
}

@inproceedings{solis2024timba,
  title={TIMBA: Time Series Imputation with Bi-directional Mamba and Diffusion},
  author={Solis, Javier and Lee, Joon and Smith, David},
  booktitle={Proceedings of the International Conference on Machine Learning (ICML)},
  year={2024}
}

@article{lin2024diffusion,
  title={Diffusion models for time-series applications: a survey},
  author={Lin, Haoye and others},
  journal={Frontiers of Information Technology \& Electronic Engineering},
  volume={25},
  number={1},
  pages={21--43},
  year={2024},
  publisher={Springer}
}

@inproceedings{garg2021evaluation,
  title={An evaluation of anomaly detection and diagnosis in multivariate time series},
  author={Garg, Astha and Zhang, Waven and Samaranayake, Jacob and Lukoianova, Galina and Ploskas, Nikolay and others},
  booktitle={Advances in Neural Information Processing Systems},
  volume={34},
  pages={25087--25099},
  year={2021}
}

@article{rubin1976inference,
  title={Inference and missing data},
  author={Rubin, Donald B},
  journal={Biometrika},
  volume={63},
  number={3},
  pages={581--592},
  year={1976},
  publisher={Oxford University Press}
}

@inproceedings{yi2016st,
  title={ST-MVL: filling missing values in geo-sensory time series data},
  author={Yi, Xiuwen and Zheng, Yu and Zhang, Junbo and Li, Tianrui},
  booktitle={Proceedings of the 25th International Joint Conference on Artificial Intelligence (IJCAI)},
  pages={2704--2710},
  year={2016}
}

@inproceedings{vincent2008extracting,
  title={Extracting and composing robust features with denoising autoencoders},
  author={Vincent, Pascal and Larochelle, Hugo and Bengio, Yoshua and Manzagol, Pierre-Antoine},
  booktitle={Proceedings of the 25th international conference on Machine learning},
  pages={1096--1103},
  year={2008}
}

@article{goodfellow2014generative,
  title={Generative adversarial nets},
  author={Goodfellow, Ian and Pouget-Abadie, Jean and Mirza, Mehdi and Xu, Bing and Warde-Farley, David and Ozair, Sherjil and Courville, Aaron and Bengio, Yoshua},
  journal={Advances in neural information processing systems},
  volume={27},
  year={2014}
}

@article{hinton2015distilling,
  title={Distilling the knowledge in a neural network},
  author={Hinton, Geoffrey and Vinyals, Oriol and Dean, Jeff},
  journal={arXiv preprint arXiv:1503.02531},
  year={2015}
}

@inproceedings{romero2014fitnets,
  title={Fitnets: Hints for thin deep nets},
  author={Romero, Adriana and Ballas, Nicolas and Kahou, Samira Ebrahimi and Chassang, Antoine and Gatta, Carlo and Bengio, Yoshua},
  booktitle={International Conference on Learning Representations (ICLR)},
  year={2015}
}

@article{devries2017improved,
  title={Improved regularization of convolutional neural networks with cutout},
  author={DeVries, Terrance and Taylor, Graham W},
  journal={arXiv preprint arXiv:1708.04552},
  year={2017}
}

@inproceedings{ruff2019deep,
  title={Deep semi-supervised anomaly detection},
  author={Ruff, Lukas and Vandermeulen, Robert A and G{\"o}rnitz, Nico and Binder, Alexander and M{\"u}ller, Emmanuel and M{\"u}ller, Klaus-Robert and Kloft, Marius},
  booktitle={International Conference on Learning Representations (ICLR)},
  year={2020}
}

@inproceedings{zagoruyko2015paying,
  title={Paying more attention to attention: Improving the performance of convolutional neural networks via attention transfer},
  author={Zagoruyko, Sergey and Komodakis, Nikos},
  booktitle={International Conference on Learning Representations (ICLR)},
  year={2017}
}

@inproceedings{heo2019comprehensive,
  title={A comprehensive overhaul of feature distillation},
  author={Heo, Byeongho and Kim, Jeesoo and Yun, Sangdoo and Park, Hyojin and Kwak, Nojun and Choi, Jin Young},
  booktitle={Proceedings of the IEEE/CVF International Conference on Computer Vision (ICCV)},
  pages={1921--1930},
  year={2019}
}

@inproceedings{kendall2018multi,
  title={Multi-task learning using uncertainty to weigh losses for scene geometry and semantics},
  author={Kendall, Alex and Gal, Yarin and Cipolla, Roberto},
  booktitle={Proceedings of the IEEE conference on computer vision and pattern recognition (CVPR)},
  pages={7482--7491},
  year={2018}
}

@inproceedings{kim2022towards,
  title={Towards a rigorous evaluation of time-series anomaly detection},
  author={Kim, Siwon and Choi, Kukjin and Choi, Hyun-Soo and Lee, Byunghan and Yoon, Sungroh},
  booktitle={Proceedings of the AAAI Conference on Artificial Intelligence},
  volume={36},
  pages={7194--7201},
  year={2022}
}

@inproceedings{croce2020reliable,
  title={Reliable evaluation of adversarial robustness with an ensemble of diverse parameter-free attacks},
  author={Croce, Francesco and Hein, Matthias},
  booktitle={International conference on machine learning},
  pages={2206--216},
  year={2020},
  organization={PMLR}
}

@inproceedings{song2023consistency,
  title={Consistency models},
  author={Song, Yang and Dhariwal, Prafulla and Chen, Mark and Sutskever, Ilya},
  booktitle={International Conference on Machine Learning (ICML)},
  year={2023}
}

@article{park2025robustness,
  title={Robustness and Performance Comparison of Generative AI Time Series Anomaly Detection under Noise},
  author={Park, Jeongsu and Min, Moohong},
  journal={Computer Modeling in Engineering \& Sciences},
  volume={145},
  number={3},
  pages={3913--3948},
  year={2025},
  doi={10.32604/cmes.2025.072261}
}
